\documentclass[lettersize,journal]{IEEEtran}
\usepackage{amsmath,amsfonts}
\usepackage{algorithmic}
\usepackage{algorithm}
\usepackage{array}
\usepackage[caption=false,font=normalsize,labelfont=sf,textfont=sf]{subfig}
\usepackage{textcomp}
\usepackage{stfloats}
\usepackage{url}
\usepackage{verbatim}
\usepackage{graphicx}
\usepackage{cite}
\usepackage[backref]{hyperref}
\usepackage{tabularray}

\begin{document}

\title{YOLIC: An Efficient Method for Object Localization and Classification on Edge Devices}

\author{\IEEEauthorblockN{Kai Su, Yoichi Tomioka, Qiangfu Zhao, Yong Liu}\\
\IEEEauthorblockA{\textit{Graduate School of Computer Science and Engineering} \\
\textit{The University of Aizu}\\
Aizuwakamatsu, Japan \\
\{d8232114, ytomioka, qf-zhao, yliu\}@u-aizu.ac.jp}

\thanks{All resources related to this study, including datasets, image annotation tool, cell designer and source code, are available to the referees at \url{https://kai3316.github.io/yolic.github.io/}}
}

\markboth{Journal of \LaTeX\ Class Files,~Vol.~14, No.~8, August~2021}%
{Shell \MakeLowercase{\textit{et al.}}: A Sample Article Using IEEEtran.cls for IEEE Journals}


\maketitle

\begin{abstract}
In the realm of Tiny AI, we introduce ``You Only Look at Interested Cells" (YOLIC), an efficient method for object localization and classification on edge devices. Through seamlessly blending the strengths of semantic segmentation and object detection, YOLIC offers superior computational efficiency and precision. By adopting Cells of Interest for classification instead of individual pixels, YOLIC encapsulates relevant information, reduces computational load, and enables rough object shape inference. Importantly, the need for bounding box regression is obviated, as YOLIC capitalizes on the predetermined cell configuration that provides information about potential object location, size, and shape. To tackle the issue of single-label classification limitations, a multi-label classification approach is applied to each cell for effectively recognizing overlapping or closely situated objects. This paper presents extensive experiments on multiple datasets to demonstrate that YOLIC achieves detection performance comparable to the state-of-the-art YOLO algorithms while surpassing in speed, exceeding 30fps on a Raspberry Pi 4B CPU.
\end{abstract}


\section{Introduction}

Pioneering algorithms such as YOLO\cite{redmon2016you}, Faster R-CNN\cite{ren2015faster}, U-Net\cite{ronneberger2015u}, and SegNet\cite{badrinarayanan2017segnet} in the realm of computer vision have made significant strides due to their outstanding performance. They have been broadly applied across multiple domains, including autonomous driving, medical image analysis, and security surveillance. However, implementing these resource-demanding methodologies on edge devices, characterized by cost, computational power, and memory constraints, poses significant challenges.

These state-of-the-art algorithms primarily address two major tasks in computer vision: object detection and image segmentation, each offering unique advantages and imposing limitations as well. Semantic segmentation model classifies every pixel within an image into a specific category, delivering a detailed understanding of the scene. While this granularity ensures high precision, it demands significant computational resources, posing feasibility challenges for deployment on resource-constrained edge devices. Conversely, object detection algorithms aim to identify and localize objects within an image by creating bounding boxes around them. These methods, being less resource-intensive than semantic segmentation, present a more practical solution for edge devices. However, their limitations become apparent when the shape information of objects is necessary, as the bounding boxes usually cannot represent the actual shape of the objects.

Several strategies have been proposed in the literature to find a balance between these methodologies. These include the use of lightweight backbone networks tailored for mobile devices\cite{sandler2018mobilenetv2}, hardware accelerators like Intel's Neural Compute Stick and Google's Coral\cite{Accelerators}, model compression techniques such as model pruning\cite{he2017channel,liu2018rethinking}, knowledge distillation\cite{wang2021knowledge}, quantization\cite{li2019fully}, and edge-cloud collaborative computing\cite{9771217}. Each of these solutions, however, involves trade-offs between accuracy, computational efficiency, and energy consumption. There is a pressing need for a solution that effectively balances these factors while accommodating the unique constraints of edge devices.

\begin{figure*}[htbp]
\centerline{\includegraphics[scale=0.5]{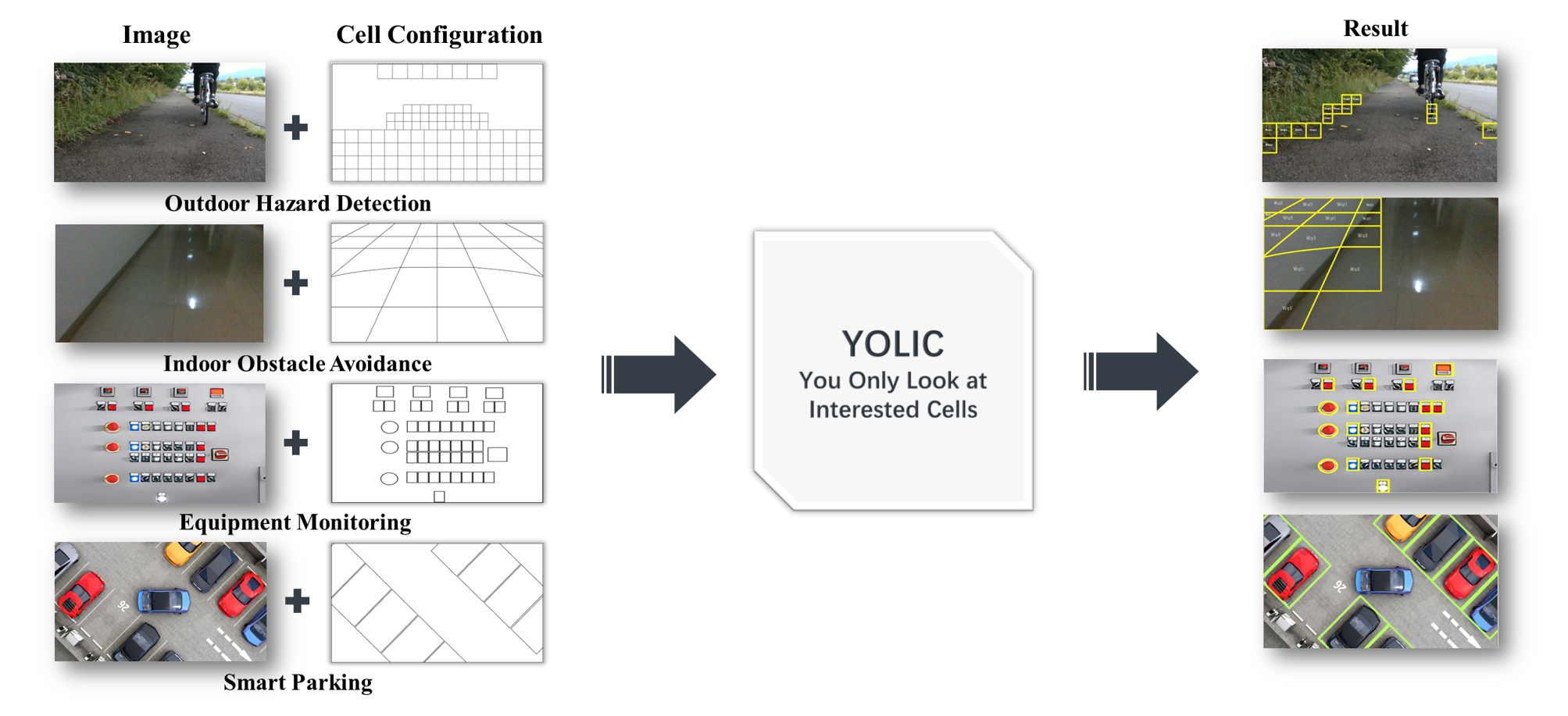}}
\caption{YOLIC's adaptability is showcased in diverse applications such as intelligent driving, industrial manufacturing, and smart parking, where it proficiently identifies various cells of interest. Unlike conventional object detection algorithms which involve laborious searching for objects in the entire image, the proposed method passively waits for the object to appear in the predefined cells of interest. This flexibility enables customized cell configurations for unique scenarios, allowing precise object detection and analysis across different tasks.}
\label{fig1}
\end{figure*}
Taking into account the challenges and limitations of edge devices, we unveil ``You Only Look at Interested Cells" (YOLIC) in Figure\ref{fig1}. This novel methodology uniquely sits at the intersection of semantic segmentation and object detection, ingeniously blending their strengths while circumventing their individual weaknesses. By doing so, YOLIC provides a more resource-conscious yet effective solution that strikes a balance between precision and computational efficiency. YOLIC employs Cell of Interest (CoI) as the unit of classification based on cell-wise segmentation rather than classifying each pixel as in semantic segmentation. These CoIs, larger than individual pixels, encapsulate more relevant information, thereby enhancing detection efficiency and reducing computational overhead. On the other hand, instead of merely enclosing objects within bounding boxes, YOLIC infers a rough shape of the objects by utilizing multiple CoIs, offering a more nuanced perspective of object presence in the scene. This is accomplished by defining a number of cells and concentrating solely on those identified as CoIs in an image. Thus computational load could be significantly reduced.

YOLIC is particularly effective in scenarios where both the shape and location of an object are crucial. Leveraging a prior knowledge of the expected object's position, size, and shape, YOLIC allows for predefined positioning and count of cells. This approach simplifies the computational process for object detection, eliminating the need for region proposal networks or bounding box regression techniques. Furthermore, YOLIC adopts multi-label classification for each cell, enhancing the detection of multiple objects within a single cell. This approach effectively recognizes overlapping or closely situated objects within the same cell, thereby overcoming the limitations associated with single-label classification. YOLIC's integration of cell-wise segmentation and multi-label classification achieves accuracy levels comparable to state-of-the-art object detection algorithms while significantly reducing computational resource demands. This balance between precision and computational efficiency positions YOLIC as an ideal solution for object detection on edge devices.

From a practical standpoint, the main advantages of YOLIC are as follows: 1) Efficiency: the fastest YOLIC model achieves real-time performance with 224 × 224 pixel inputs at 40.06 FPS on a Raspberry Pi 4B CPU, making it suitable for resource-limited environments and real-time applications. 2) Accuracy: YOLIC demonstrates excellent detection performance across diverse datasets and cell configurations, highlighting its versatility and effectiveness in various object localization/classification tasks. 3) Compatibility: YOLIC can be easily adapted from existing classification models with minimal modifications. 4) Cost-Effectiveness: The successful implementation of YOLIC on affordable edge devices like the Raspberry Pi CPU showcases the potential for cost-effective object detection solutions. 5) Pioneering: To the best of our knowledge, YOLIC is the first method with 224 × 224 pixel inputs to surpass 30 FPS detection on a Raspberry CPU (Broadcom BCM2711), delivering accuracy comparable to cutting-edge YOLO variants. This unprecedented speed and precision open up new possibilities for Internet of Things (IoT) applications, including integration into road obstacle detection systems, security cameras, and a wide range of IoT devices.

The structure of this paper is as follows. In Section II, we provide a overview of current object detection techniques on resource-constrained devices and other relevant techniques. In Chapter III, we introduce the preliminaries, detailing the foundational aspects. In Section IV, we introduce YOLIC, detailing its distinctive features. In Section V, we present a range of the experiments and result analysis, followed by the conclusion in Chapter VI.

\section{Related Work}
\subsection{Object Detection on Resource-Constrained Devices}
One-stage object detection techniques have emerged as the method of choice for real-time object detection, owing to their unified detection pipeline and impressive performance on general-purpose GPU devices.  Prominent one-stage algorithms, such as YOLO\cite{redmon2016you}, EfficientDet\cite{tan2020efficientdet}, RetinaNet\cite{lin2017focal}, and SSD\cite{liu2016ssd}, streamline the detection process by eliminating the need for a separate region proposal stage, which is often used in two-stage solutions like R-CNN\cite{girshick2014rich} and its derivatives\cite{girshick2015fast, ren2015faster}. Nevertheless, current one-stage object detection methodologies may not be sufficiently suited for deployment across a broad spectrum of resource-constrained devices. This challenge can be primarily ascribed to two factors: 1) Performance on Low-Cost Devices: contemporary state-of-the-art lightweight algorithms for object detection, such as NanoDet\cite{nanodet} and YOLOX-Nano\cite{ge2021yolox}, have been specifically tailored for resource-limited devices. Although these algorithms have demonstrated successful real-time object detection capabilities with a running speed of 30fps+ on high-end mobile chipsets indicated in article\cite{yu2021pp}, their performance on low-cost hardware platforms, like the Raspberry Pi, leaves much to be desired. For example, the frame rate of these models on such devices is typically around 10, which falls significantly short of the ideal real-time performance. 2) Detection Accuracy Gap: To achieve a lower computational cost, lightweight models often sacrifice detection accuracy. This trade-off is evident when comparing the performance of standard and lightweight models on benchmark datasets. For instance, YOLOX-M attains an average precision of 46.4 on the COCO object detection dataset, while YOLOX-Nano achieves an average precision of only 25.3, resulting in a substantial gap of approximately 1.8 times\cite{ge2021yolox}. In summary, despite the tremendous potential of one-stage object detection techniques for real-time detection on general-purpose GPU devices, there remains substantial room for improvement in terms of their performance on resource-constrained hardware. 
\subsection{Cell-wise Semantic Segmentation}
Cell-wise semantic segmentation is an innovative image processing method that merges the strengths of semantic segmentation and object detection, producing a more precise strategy for object localization within an image. It works by breaking down an image into smaller segments or cells, each scrutinized independently for the presence of object components. An example of this approach can be seen in the one-image version of Google ReCAPTCHA\cite{recaptcha}, where an image is divided into a 16-grid layout, and the user is tasked with identifying all cells containing a specific object. In our prior research\cite{9251174}, we applied this concept for road risk detection. We divided the road region of an image into numerous sub-images. Each of these sub-images were then combined into a single batch and fed into a convolutional neural network (CNN) for classification. This allowed us to identify and localize obstacles on the road within all the cells simultaneously, using just one network. However, an issue lies in its treatment of each cell as an independent unit during classification, disregarding potential relationships between cells. If the cells are too small, they may not contain enough information for reliable road risk detection. Conversely, too large cells can slow down inference speed and reduce the precision of object localization. 

\section{Preliminaries}
To make this paper relatively self-contained, we introduce multi-label classification and two CNN models suitable for mobile devices. They are helpful for constructing the model proposed in this study.
\subsection{Multi-label Classification Approach}
Multi-label classification\cite{tsoumakas2007multi} is a approach where an instance object can be classified into multiple categories simultaneously, differing from multi-class classification where each instance is assigned to only one category. Multi-label classification becomes particularly useful in scenarios where the categories are not mutually exclusive, meaning that an object can belong to more than one category. In the field of object detection, multi-label classification assumes significant relevance. Traditionally, most object detection algorithms were designed to tackle multi-class classification problems, assuming each object to belong to only one category. However, this approach might be insufficient in real-world scenarios where objects can belong to multiple categories concurrently. Beginning with YOLOv3\cite{yolov3}, some object detection algorithms started incorporating multi-label classification techniques. In these algorithms, softmax is not used for class probabilities. Instead, each class score is predicted using logistic regression, and a threshold is applied to predict multiple labels for an object. The main application of multi-label classification in these algorithms, however, is confined only within each bounding box to facilitate classification.  Our proposed method utilizes multi-label classification to identify multiple cells in an image, where each cell can be classified into multiple classes concurrently. Moreover, research\cite{GONG2019174} shows that multi-label classification can improve the accuracy of object detection when used as an auxiliary task. This further underscores the potential benefits of integrating multi-label classification into our detection frameworks.

\subsection{Approaches to Optimizing CNN for Mobile Devices}
The rise of mobile technology has led to an increasing demand for efficient and effective CNN that can operate on mobile devices. These specially designed CNNs are constructed to meet the unique constraints of mobile platforms, which include limited computational capacity, memory, and energy resources. To alleviate the computational and storage burden of CNN, techniques such as network pruning\cite{lin2020improved} (removing unnecessary connections), quantization\cite{yang2019quantization} (reducing the precision of the network's weights) and edge-cloud collaborative computing\cite{liu2021collaborative} (leveraging both edge and cloud resources for computation) have been employed. In past few years, innovative CNN architectures that offer new optimization techniques have been developed. Notably, MobileNet V2\cite{sandler2018mobilenetv2} and ShuffleNet V2\cite{ma2018shufflenet} are two prominent examples of such innovative CNN architectures. MobileNet V2 introduces two novel operations known as the inverted residual structure and linear bottleneck. These operations aim to reduce computation cost while maintaining the network's representational capacity, effectively balancing efficiency and accuracy, and making it ideal for mobile devices requiring real-time processing. On the other hand, ShuffleNet V2, aimed at improving speed and efficiency without sacrificing performance, incorporates a novel channel shuffle operation and pointwise group convolution to cut computation costs. The utilization of these lightweight architectures as the backbone network to escalate inference speed in detection tasks has become a commonplace practice\cite{9354012,10042233}.

\section{Proposed Method}
``You Only Look at Interested Cells" (YOLIC)\footnote{The foundational concept presented in this paper has been previously introduced in the literature \cite{9319469}. However, it is noteworthy that the referenced work employed a significantly more limited approach, utilizing only 22 CoIs, whereas our present work enhances this methodology, implementing a more extensive and flexible use of CoIs.} is specifically designed for edge devices and low-performance hardware. Rather than merely classifying predefined pixels of a compressed image, YOLIC utilizes a more complex process, identifying object categories by classifying CoIs of various shapes and sizes, based on the application's requirements. This approach effectively converts each input image into a `mosaic art image,' composed of numerous diverse CoIs.
The cell configuration, defining the size, shape, position, and quantity of CoIs, is what lends YOLIC its precision and versatility in object localization, differing from traditional algorithms that rely primarily on object coordinates. These cells, tailored according to factors such as distance, direction, and the degree of danger posed by frequently appearing objects, provide sufficient and critical information for tasks like safe driving.
Figure \ref{fig1} demonstrates the versatility of YOLIC in detecting a wide range of CoIs for different tasks. For instance, in intelligent driving applications, CoIs can be set according to driving distance for accurate detection of the distance between objects and the vehicle. In industrial manufacturing scenarios, CoIs can be configured based on equipment panels for monitoring the operational status of traditional industrial machinery. Additionally, in smart parking lot applications, CoIs can be tailored to individual parking spaces to precisely locate vehicles. Furthermore, YOLIC's adaptability extends to various other applications such as retail analytics, where CoIs can be set to monitor product shelf areas for inventory management and customer behavior analysis. In healthcare, CoIs can be adjusted to track and analyze specific body parts or regions in medical imaging for early diagnosis and treatment planning. Overall, YOLIC's flexibility in adjusting CoIs across a multitude of tasks showcases its potential for widespread adoption in diverse industries and use cases.

\begin{figure*}[htbp]
\centerline{\includegraphics[scale=0.5]{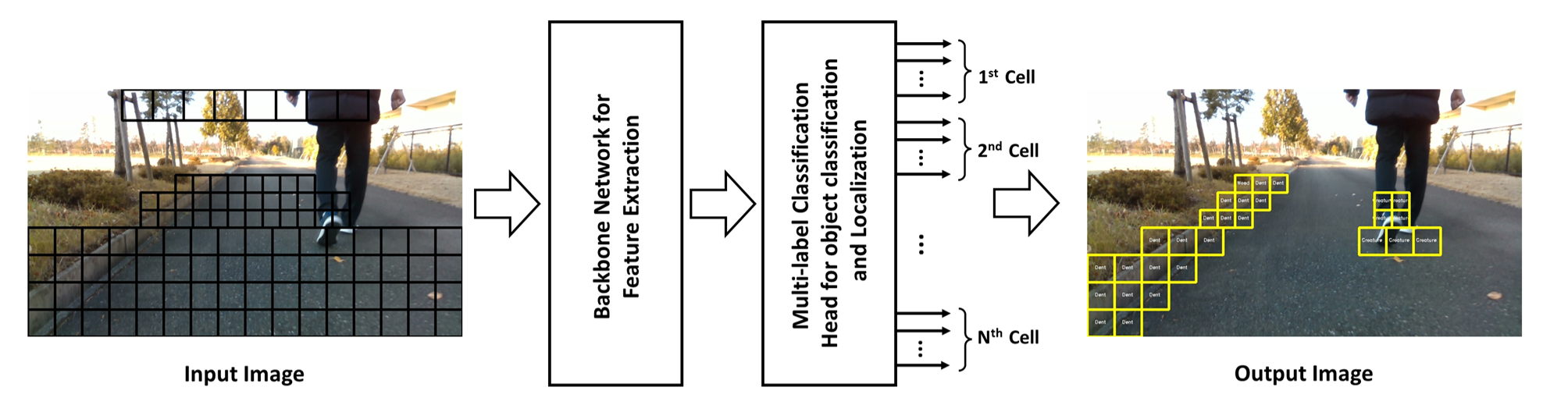}}
\caption{A detailed diagram of the YOLIC model, showcasing its two main components: the feature extraction module and the multi-label classification head. The figure also demonstrates the output of the model, highlighting the position of each CoI on an input image. The overall structure emphasizes YOLIC's lightweight and efficient design, which makes it well-suited for edge devices with limited computational resources.}
\label{fig2}
\end{figure*}

\subsection{Network Design}
The YOLIC model comprises two primary modules: a feature extraction module and a multi-label classification head in Figure\ref{fig2}. The feature extraction module primarily leverages the architecture of ShuffleNet V2, known for its efficiency, particularly for devices with limited computational resources. The feature map extracted from this module is subsequently input to the multi-label classification head, sequentially outputting the classification results for each CoI. In our model, the input size is set to be 224. This is because YOLIC focuses on detecting objects within predefined CoI, it does not require high-resolution images to accurately localize objects. The chosen input size ensures a balance between computational efficiency and feature detail preservation. The multi-label classification head comprises three components: an adaptive average pooling layer, a Dropout layer and a fully connected layer, outputting a distinct prediction for each CoI. The classifier's total number of outputs $C$ is calculated as follow:
\begin{equation} \label{eq1}
C = N \times (M + 1)
\end{equation}
where $N$ represents the number of CoIs, $M+1$ stands for the number of objects of interest within each CoI plus one background class. The inclusion of the background class helps the model distinguish between cells containing objects of interest and those without, thereby improving the binary detection performance for each CoI.

One of the defining characteristics of the YOLIC model is its inherent flexibility concerning the feature extraction module. While ShuffleNet V2 is employed in the initial design, this module can be substituted with alternative networks to balance detection speed and performance. We can also use MobileNet V2 as YOLIC's feature extraction module. This variant of YOLIC resulted in a model with a high level of detection performance, albeit at a slightly slower detection speed.

The overall diagram of the YOLIC model, including both the feature extraction module and the multi-label classification head, is depicted in Figure \ref{fig2}, providing a comprehensive understanding of the YOLIC's concept. Table \ref{tm} detailed the architecture of the YOLIC when employing ShuffleNet V2.

\begin{table}[htbp]
\centering
\caption{This table delineates the architecture of the YOLIC model using ShuffleNet V2. It details the layer structure of the feature extraction module and the multi-label classification head, which include output size, filter size, output channels, and number of repeated blocks (B). C is the value given in Eq.(\ref{eq1})}
\label{tm}
\resizebox{\linewidth}{!}{%
\begin{tblr}{
		cells = {c},
		cell{2}{1} = {r=7}{},
		cell{9}{1} = {r=3}{},
		hline{1-2,9,12} = {-}{},
	}
	Module                                & Layer                  & Output Size & Filter & output channels & B \\
	{Feature\\~Extraction\\~Module}       & Input Layer            & 224x224x3   & -      & 3               & 1 \\
	& Standard Convolution   & 112x112x24  & 3x3    & 24              & 1 \\
	& MaxPool                & 56x56x24    & 3x3    & 24              & 1 \\
	& ShuffleUnit            & 28x28x116   & -      & 116             & 4 \\
	& ShuffleUnit            & 14x14x116   & -      & 232             & 8 \\
	& ShuffleUnit            & 7x7x116     & -      & 464             & 4 \\
	& Pointwise Convolution  & 7x7x1024    & 1x1    & 1024            & 1 \\
	{Multi-Label \\Classification\\~Head} & Global Average Pooling & 1x1x1024    & -      & 1024            & - \\
	& Dropout Layer          & 1x1x1024    & -      & 1024            & - \\
	& Fully Connected Layer  & $C$           & -      & -              & - 
\end{tblr}
}
\end{table}

\subsection{Design of CoI Configuration}
YOLIC presents a distinctive approach compared to existing object detection algorithms, as it exclusively concentrates on objects within predefined CoIs. This approach highlights the principle of selective attention, significantly boosting its efficiency and adaptability across diverse detection scenarios. The cell configuration is not fixed but can be customized to meet specific detection requirements. Different scenarios may necessitate varying cell configurations to effectively capture and classify objects of interest. To streamline this process, we have developed a cell designer tool\cite{Su_Cell_Designer_Tool_2023} that simplifies the task of defining and partitioning cells within images.

Once the CoI configuration is established, image annotations adapt based on this setup and the objects of interest. The resultant annotation files only contain the class label for each CoI, eliminating coordinate information. This approach offers a straightforward representation of objects within CoIs, streamlines the network's learning process, reduces annotation efforts, and lessens training complexity. However, this unique annotation method presents challenges, specifically the need for an efficient way to annotate different CoIs within varied cell configurations. To tackle this, we have developed an annotation tool\cite{Su_Image_Annotation_Tool_2023} that supports various cell configurations, designed with the cell designer tool, and promotes efficient annotation of the CoIs. This tool expedites dataset preparation for YOLIC, minimizing time and effort.

\subsection{Inference}
YOLIC's inference process focuses on simplicity and efficiency, which further bolsters its applicability to a wide range of tasks. During inference, YOLIC outputs the class probabilities for all CoIs as determined by the cell configuration.

The model's output is organized according to the CoIs' coordinates specified in the cell configuration, allowing for a direct mapping between the output and the image's spatial locations. By examining the background bit for each CoI, we can determine whether a class of interest is present within that cell. Furthermore, for each class of interest, we can obtain the probability that a particular object appears within the CoIs. When both the probabilities of the background class and the object class exceed a set threshold, typically defined as 0.5, the model primarily relies on the background class's result to conclude the presence of an object.

An important advantage of YOLIC's inference process is the absence of time-consuming and computationally intensive post-processing algorithms, such as non-maximum suppression (NMS), commonly used in traditional object detection methods. This streamlined approach contributes to YOLIC's suitability for resource-constrained environments and real-time applications, while still delivering accurate object detection results.

\subsection{Training}
We utilize transfer learning in our experiments to improve performance and efficiency. We maintain the pre-trained weights obtained from ImageNet dataset in the feature extraction component, since these weights offer a solid initialization for the model. In this proposed framework, the network's multi-label classification head predicts the class probabilities for all CoIs, as well as for each individual object contained within these CoIs. To ensure the independence of the outputs, a Sigmoid activation function is integrated into the classification head. During the training phase, a binary cross-entropy loss function is employed as:

\begin{equation}
	\resizebox{.9\hsize}{!}{$
		Loss = -\frac{1}{C} \sum_{i=1}^{N}{ \sum_{j=1}^{M+1}[y_{ij}log(p_{ij})+(1-y_{ij})log(1-p_{ij})]}
		$}
\end{equation}
where $C$ is given in Eq.(\ref{eq1}), $y_{ij}$ denotes the true label for the j-th object in the i-th CoI, $p_{ij}$ represents the predicted probability for the j-th object in the i-th CoI.

To further enhance YOLIC’s suitability for edge devices, quantization-aware training is used to obtain a set of quantized models. Combining the YOLIC network design and quantization-aware training enables us to optimize model performance while significantly reducing computational requirements. This results in lightweight, power-efficient models that maintain robust detection capabilities, ideal for deployment on resource-constrained edge devices.

\section{Experiments and Evaluation}
In this section, we conduct experiments to demonstrate YOLIC's adaptability across varied scenarios. Our primary focus is on two environments: outdoor hazard detection, using a curated dataset with regular rectangular cells, and indoor obstacle avoidance, leveraging a dataset with irregularly shaped cells. This adaptability to irregular cells underscores YOLIC's unique strength, facilitating effective object detection in complex spatial configurations, particularly indoors. Thus, YOLIC's flexible design enhances its precision and utility across diverse detection scenarios. To supplement these primary experiments and validate the viability of our proposed method on one public dataset, we utilized the Cityscapes semantic segmentation dataset\cite{Cordts_2016_CVPR}, owing to its rich pixel-level annotations covering a broad range of object categories. We converted their pixel-wise labeled images to fit our selected objects, adapting them to the YOLIC object detection approach based on our designed cell configurations. We have made all these datasets publicly accessible on our project website at \url{https://kai3316.github.io/yolic.github.io/}, facilitating the reproduction of our results.

Our experiments were executed using the PyTorch framework on an NVIDIA RTX 4090 GPU. We leveraged the Adam optimizer with an initial learning rate of 0.001 and a MultiStepLR scheduler with milestones at 100th and 125th epochs for learning rate adjustment. All YOLIC model was trained with input images of 224$\times$224 pixels and a batch size of 32 across 150 epochs. Data augmentation techniques, including random horizontal flipping and color jittering, were utilized for model robustness.

In training the comparative models YOLOv5, YOLOv6, and YOLOv8, we accounted for YOLIC's positional-dependent characteristics and made necessary adjustments. To ensure fairness, the same ground truth used in YOLIC were also applied to these models. The mosaic augmentation, which could alter the original layout of training images, was disabled. Despite these changes, all models kept their original structures and input image sizes. The training was conducted over a maximum of 300 epochs to ensure that these comparative models received sufficient training, with automatic framework settings for the optimizer and batch size, and early stopping is employed to avoid overfitting. 

\begin{figure}[htbp]
	\centerline{\includegraphics[scale=0.37]{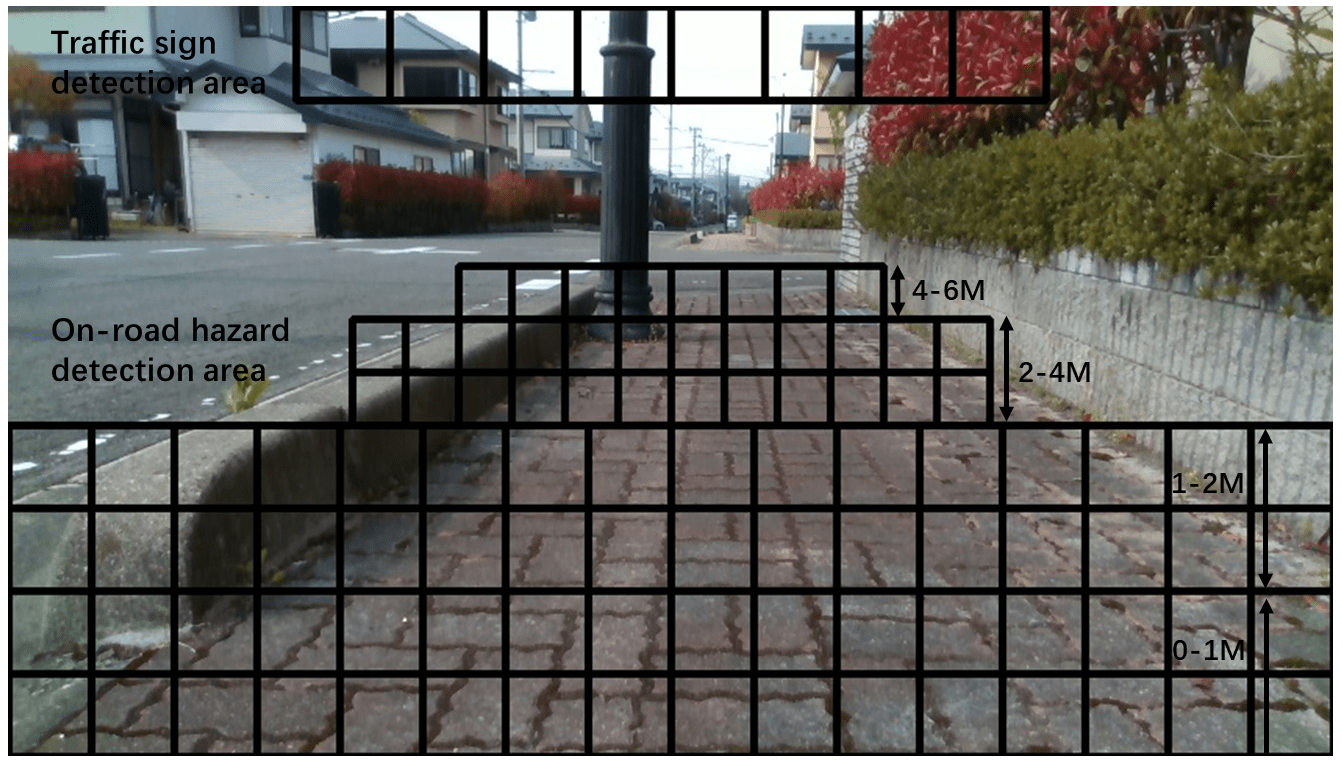}}
	\caption{Cell configuration designed for outdoor risk detection on electric scooters, illustrating the distribution of 96 CoIs for road hazards within a 0-6 meter range and eight additional CoIs for traffic sign localization.}
	\label{fig5}
\end{figure}

\subsection{Outdoor Hazard Detection}
In this experiment, we focus on the development of an outdoor risk detection system specifically tailored for low-cost electric scooters. The motivation behind using YOLIC for this application lies in its ability to provide efficient real-time detection, which is crucial for ensuring safety during scooter navigation. Furthermore, low-cost electric scooters typically have limited computational resources, making YOLIC's lightweight architecture an ideal choice for deployment on such devices.

To accommodate the typical speed of electric scooters and ensure safety, we designed a detection range of 0-6 meters on the road. We divided this range into different CoIs based on distance, allowing for precise object localization and prioritization of areas posing higher risks. In total, 96 CoIs are allocated to the road detection area. Additionally, we designated a traffic sign detection area at the top of the image, employing eight CoIs to approximate the location of traffic signs. The 104-cell configuration for this experiment is illustrated in Figure \ref{fig5}.

\begin{figure*}[htbp]
	\centerline{\includegraphics[scale=0.53]{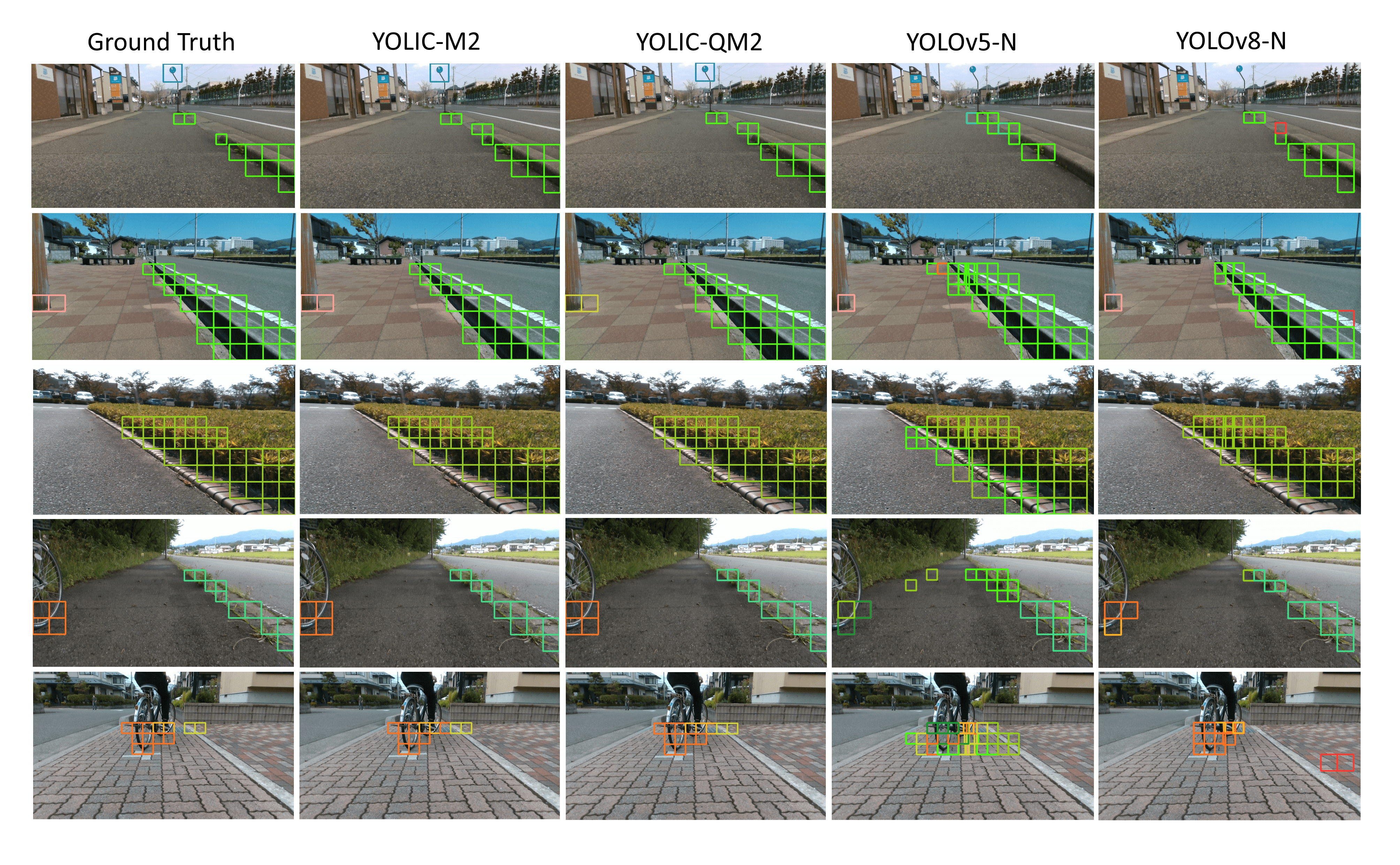}}
	\caption{We compare the detection results of our proposed models with the YOLO by visualizing the output on sample outdoor images. The four models with the highest F1-scores are selected for the comparison.}
	\label{outdoor}
\end{figure*}

\begin{table*}
	\caption{Detailed comparison of outdoor hazard detection performance for various objects of interest. Precision and recall scores of each model are presented.}
	\label{t3}
	\resizebox{\linewidth}{!}{%
		\begin{tblr}{
				cells = {c},
				cell{1}{1} = {r=2}{},
				cell{1}{2} = {r=2}{},
				cell{1}{3} = {c=3}{},
				cell{1}{6} = {c=3}{},
				cell{1}{9} = {c=3}{},
				cell{1}{12} = {c=3}{},
				cell{1}{15} = {c=3}{},
				cell{1}{18} = {c=3}{},
				cell{13}{1} = {r=2}{},
				cell{13}{2} = {r=2}{},
				cell{13}{3} = {c=3}{},
				cell{13}{6} = {c=3}{},
				cell{13}{9} = {c=3}{},
				cell{13}{12} = {c=3}{},
				cell{13}{15} = {c=3}{},
				cell{13}{18} = {c=3}{},
				hline{1,3,12-13,15,24} = {-}{},
			}
			Method    & Input Size & Bump      &        &          & Column    &        &          & Dent          &        &          & Fence       &        &          & People    &        &          & Vehicle   &        &          \\
			&            & Precision & Recall & F1-score & Precision & Recall & F1-score & Precision     & Recall & F1-score & Precision   & Recall & F1-score & Precision   & Recall & F1-score & Precision & Recall & F1-score \\
			YOLOv5-N  & 640×640    & 0.4540    & 0.5600 & 0.5015   & 0.2410    & 0.2670 & 0.2533   & 0.8130        & 0.1580 & 0.2646   & 0.0937      & 0.3040 & 0.1432   & 0.8070      & 0.5900 & 0.6816   & 0.5390    & 0.6260 & 0.5793   \\
			YOLOv5-S  & 640×640    & 0.3600    & 0.6780 & 0.4703   & 0.1470    & 0.5640 & 0.2332   & 0.4230        & 0.0410 & 0.0748   & 0.1170      & 0.1320 & 0.1240   & 0.8220      & 0.4140 & 0.5507   & 0.5570    & 0.6950 & 0.6184   \\
			YOLOv6-N  & 640×640    & 0.5990    & 0.4600 & 0.5204   & 0.0875    & 0.4200 & 0.1448   & 0.7000        & 0.2300 & 0.3462   & 0.1220      & 0.4150 & 0.1886   & 0.4280      & 0.5810 & 0.4929   & 0.3830    & 0.3840 & 0.3835   \\
			YOLOv8-N  & 640×640    & 0.8670    & 0.7930 & 0.8284   & 0.6620    & 0.7180 & 0.6889   & 0.8720        & 0.6810 & 0.7648   & 0.8310      & 0.3710 & 0.5130   & 0.8920      & 0.8500 & 0.8705   & 0.8720    & 0.8790 & 0.8755   \\
			YOLOv8-S  & 640×640    & 0.3640    & 0.8640 & 0.5122   & 0.2990    & 0.4730 & 0.3664   & 0.7600        & 0.4080 & 0.5310   & 0.0598      & 0.0990 & 0.0746   & 0.8570      & 0.5880 & 0.6975   & 0.7750    & 0.6610 & 0.7135   \\
			YOLIC-M2  & 224×224    & 0.9147    & 0.8897 & 0.9020   & 0.7998    & 0.7689 & 0.7840   & 0.8940        & 0.8498 & 0.8713   & 0.8900      & 0.8242 & 0.8558   & 0.8457      & 0.8152 & 0.8302   & 0.8975    & 0.9010 & 0.8992   \\
			YOLIC-S2  & 224×224    & 0.8966    & 0.8745 & 0.8854   & 0.7830    & 0.7065 & 0.7428   & 0.8739        & 0.8496 & 0.8616   & 0.8484      & 0.8213 & 0.8346   & 0.8039      & 0.7913 & 0.7976   & 0.8846    & 0.8807 & 0.8826   \\
			YOLIC-QM2 & 224×224    & 0.9192    & 0.8642 & 0.8909   & 0.8148    & 0.7130 & 0.7605   & 0.8926        & 0.8306 & 0.8605   & 0.9025      & 0.7996 & 0.8479   & 0.8502      & 0.7817 & 0.8145   & 0.8977    & 0.8794 & 0.8885   \\
			YOLIC-QS2 & 224×224    & 0.9281    & 0.8090 & 0.8645   & 0.8364    & 0.6049 & 0.7021   & 0.9194        & 0.7379 & 0.8187   & 0.8903      & 0.7600 & 0.8200   & 0.8556      & 0.7023 & 0.7714   & 0.9128    & 0.8110 & 0.8589   \\
			&            &           &        &          &           &        &          &               &        &          &             &        &          &             &        &          &           &        &          \\
			Method    & Input Size & Wall      &        &          & Weed      &        &          & ZebraCrossing &        &          & TrafficCone &        &          & TrafficSign &        &          & All       &        &          \\
			&            & Precision & Recall & F1-score & Precision & Recall & F1-score & Precision     & Recall & F1-score & Precision   & Recall & F1-score & Precision   & Recall & F1-score & Precision & Recall & F1-score \\
			YOLOv5-N  & 640×640    & 0.3650    & 0.4060 & 0.3844   & 0.6370    & 0.7860 & 0.7037   & 0.8420        & 0.1470 & 0.2503   & 0.0948      & 0.3070 & 0.1449   & 0.0000      & 0.0000 & 0.0000   & 0.4442    & 0.3774 & 0.4081   \\
			YOLOv5-S  & 640×640    & 0.2130    & 0.7540 & 0.3322   & 0.5630    & 0.9230 & 0.6994   & 0.5480        & 0.5370 & 0.5424   & 0.0148      & 0.0151 & 0.0149   & 0.0000      & 0.0000 & 0.0000   & 0.3423    & 0.4321 & 0.3820   \\
			YOLOv6-N  & 640×640    & 0.3250    & 0.4400 & 0.3739   & 0.4080    & 0.4310 & 0.4192   & 0.7850        & 0.5730 & 0.6625   & 0.1380      & 0.7260 & 0.2319   & 0.2710      & 0.0186 & 0.0348   & 0.3860    & 0.4253 & 0.4047   \\
			YOLOv8-N  & 640×640    & 0.8390    & 0.8330 & 0.8360   & 0.8410    & 0.8980 & 0.8686   & 0.8820        & 0.8080 & 0.8434   & 0.7800      & 0.6300 & 0.6970   & 1.0000      & 0.0183 & 0.0359   & 0.8489    & 0.6799 & 0.7551   \\
			YOLOv8-S  & 640×640    & 0.6600    & 0.7160 & 0.6869   & 0.6360    & 0.9230 & 0.7531   & 0.7820        & 0.5810 & 0.6667   & 0.2280      & 0.1550 & 0.1845   & 0.0000      & 0.0000 & 0.0000   & 0.4928    & 0.4971 & 0.4949   \\
			YOLIC-M2  & 224×224    & 0.9367    & 0.9215 & 0.9290   & 0.9307    & 0.9182 & 0.9244   & 0.9728        & 0.9575 & 0.9651   & 0.8507      & 0.7668 & 0.8066   & 0.8453      & 0.7116 & 0.7727   & 0.8889    & 0.8477 & 0.8678   \\
			YOLIC-S2  & 224×224    & 0.9265    & 0.9131 & 0.9198   & 0.9179    & 0.9144 & 0.9161   & 0.9769        & 0.9529 & 0.9648   & 0.8102      & 0.7529 & 0.7805   & 0.8287      & 0.6977 & 0.7576   & 0.8682    & 0.8323 & 0.8499   \\
			YOLIC-QM2 & 224×224    & 0.9365    & 0.9123 & 0.9242   & 0.9326    & 0.9075 & 0.9199   & 0.9732        & 0.9478 & 0.9603   & 0.8705      & 0.7251 & 0.7912   & 0.8427      & 0.6977 & 0.7634   & 0.8939    & 0.8235 & 0.8573   \\
			YOLIC-QS2 & 224×224    & 0.9456    & 0.8762 & 0.9096   & 0.9442    & 0.8754 & 0.9085   & 0.9837        & 0.9329 & 0.9576   & 0.8533      & 0.6949 & 0.7660   & 0.8980      & 0.6140 & 0.7293   & 0.9061    & 0.7653 & 0.8298   
		\end{tblr}
	}
\end{table*}

\begin{table*}
	\centering
	\caption{Binary classification performance of the proposed YOLIC models for outdoor hazard detection. }
	\label{t4}
		\begin{tblr}{
				cells = {c},
				cell{1}{1} = {r=2}{},
				cell{1}{2} = {r=2}{},
				cell{1}{3} = {c=3}{},
				cell{1}{6} = {c=3}{},
				cell{1}{9} = {c=3}{},
				hline{1,3,7} = {-}{},
			}
			Method    & Input Size & Risk      &        &          & Road      &        &          & All       &        &          \\
			&            & Precision & Recall & F1-score & Precision & Recall & F1-score & Precision & Recall & F1-score \\
			YOLIC-M2  & 224 × 224        & 0.9426    & 0.9410 & 0.9418   & 0.9876    & 0.9880 & 0.9878   & 0.9651    & 0.9645 & 0.9648   \\
			YOLIC-S2  & 224 × 224        & 0.9311    & 0.9354 & 0.9332   & 0.9864    & 0.9855 & 0.9859   & 0.9588    & 0.9605 & 0.9596   \\
			YOLIC-QM2 & 224 × 224        & 0.9252    & 0.9475 & 0.9362   & 0.9889    & 0.9839 & 0.9864   & 0.9571    & 0.9657 & 0.9614   \\
			YOLIC-QS2 & 224 × 224        & 0.8941    & 0.9556 & 0.9238   & 0.9905    & 0.9763 & 0.9833   & 0.9423    & 0.9660 & 0.9540   
		\end{tblr}
\end{table*}

For this application, it is essential to accurately identify various objects that could potentially pose risks to scooter navigation. Accordingly, we defined 11 objects of interest: Bump, Column, Dent, Fence, People, Vehicle, Wall, Weed, Zebra Crossing, Traffic Cone, and Traffic Sign. These objects were selected due to their relevance and potential impact on scooter movement and safety. Therefore, considering the 11 objects and one additional category for background class, YOLIC's output for this experiment becomes $104 \times (11+1) = 1248$.

In our evaluation, we hand-picked and annotated 20,380 distinct video frames from road footage around the University of Aizu campus. These frames were carefully chosen to provide unique data points. For the experiment, we randomly divided this dataset into 70\% for training, 10\% for validation, and the remaining 20\% for testing. This dataset showcasing varied outdoor conditions, facilitates a comprehensive assessment of the effectiveness of our proposed YOLIC-based risk detection system for electric scooters. We evaluate our model using common metrics such as precision, recall, and the F1-score for each object category. We also appraise the binary classification of each CoI based on the results of the background class.

Table \ref{t3} provides a comprehensive comparison of our proposed models, where `S2' and `M2' denote the use of ShuffleNet V2 and MobileNet V2 for feature extraction, respectively. `QS2' and `QM2' refer to the quantized versions of these models, a process which simplifies the model by reducing the precision of its weights, thereby increasing computational efficiency without a significant drop in performance. From the comparison, it is clear that our proposed models consistently outperform the YOLO series, including YOLOv5\cite{Jocher_YOLOv5}, YOLOv6\cite{li2022yolov6}, and YOLOv8\cite{Jocher_YOLOv8} in terms of precision and recall. Notably, YOLIC-M2 exhibits high precision and recall across all categories, achieving an impressive F1-score of 0.8678 for the All category. Moreover, YOLIC-S2, YOLIC-QM2, and YOLIC-QS2 demonstrate excellent performance, further attesting to our methodology's robustness. Table \ref{t4} presents the binary classification results for each CoI, showcasing our models' effectiveness in distinguishing between `Risk' and `Road' categories. All YOLIC models displayed similar high performance with strong F1-scores. Figure \ref{outdoor} provides a visual comparison of our models with the YOLO series.

\subsection{Indoor Obstacle Avoidance}
In this experiment, our primary focus is on detecting ground-level obstacles, a key aspect in ensuring the safe operation of autonomous vehicles or robots in indoor environments. Traditional object detection techniques mainly use grid-based bounding box strategies, which often fail to adequately accommodate the complex and varied characteristics of indoor landscapes.

\begin{figure}[htbp]
	\centerline{\includegraphics[scale=0.36]{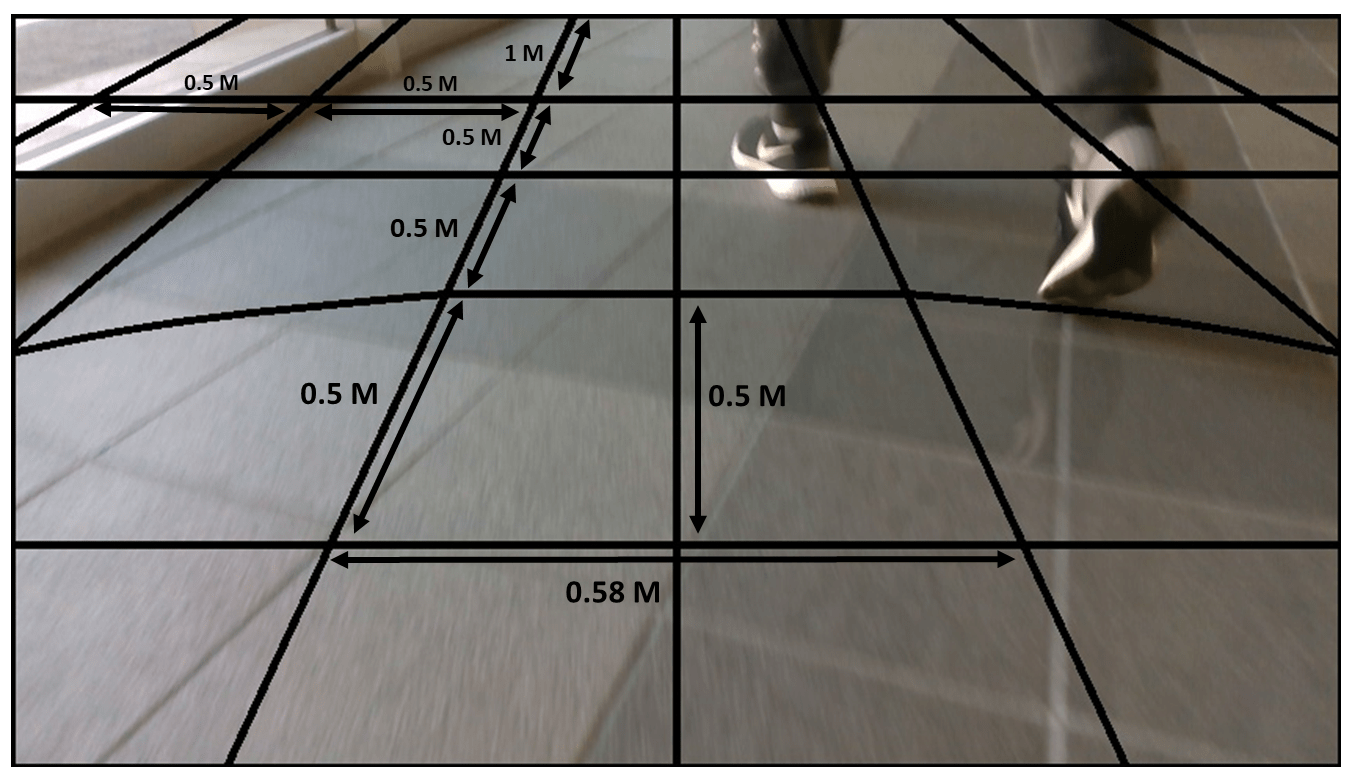}}
	\caption{Cell configuration for indoor obstacle detection experiment, showcasing 30 irregularly-shaped CoIs distributed across the video frame.}
	\label{fig6}
\end{figure}

In light of this, we have adopted a distinctive cell configuration, utilizing irregularly-shaped CoIs, intending to enhance the precision of obstacle localization. As depicted in Figure \ref{fig6}, this cellular layout incorporates 30 irregularly-shaped CoIs dispersed throughout the entire video frame. Each CoI is defined based on distance estimations, consequently enabling a more useful information for autonomous driving. 

Specifically, the vehicle employed in our experiment has a width of 0.58 meters. Based on this width, we have located ten CoIs directly in front of the vehicle to detect potential front-facing obstacles. Moreover, on both the left and right flanks of the vehicle, we have positioned two quarter-circle-shaped CoIs with a 1-meter radius. These quarter-circle-shaped CoIs are specifically designed to detect objects within one meter distance that could be obstacles when making left or right turns, a crucial aspect for the vehicle's safe maneuvering and collision avoidance in confined spaces.

\begin{figure*}[htbp]
	\centerline{\includegraphics[scale=0.52]{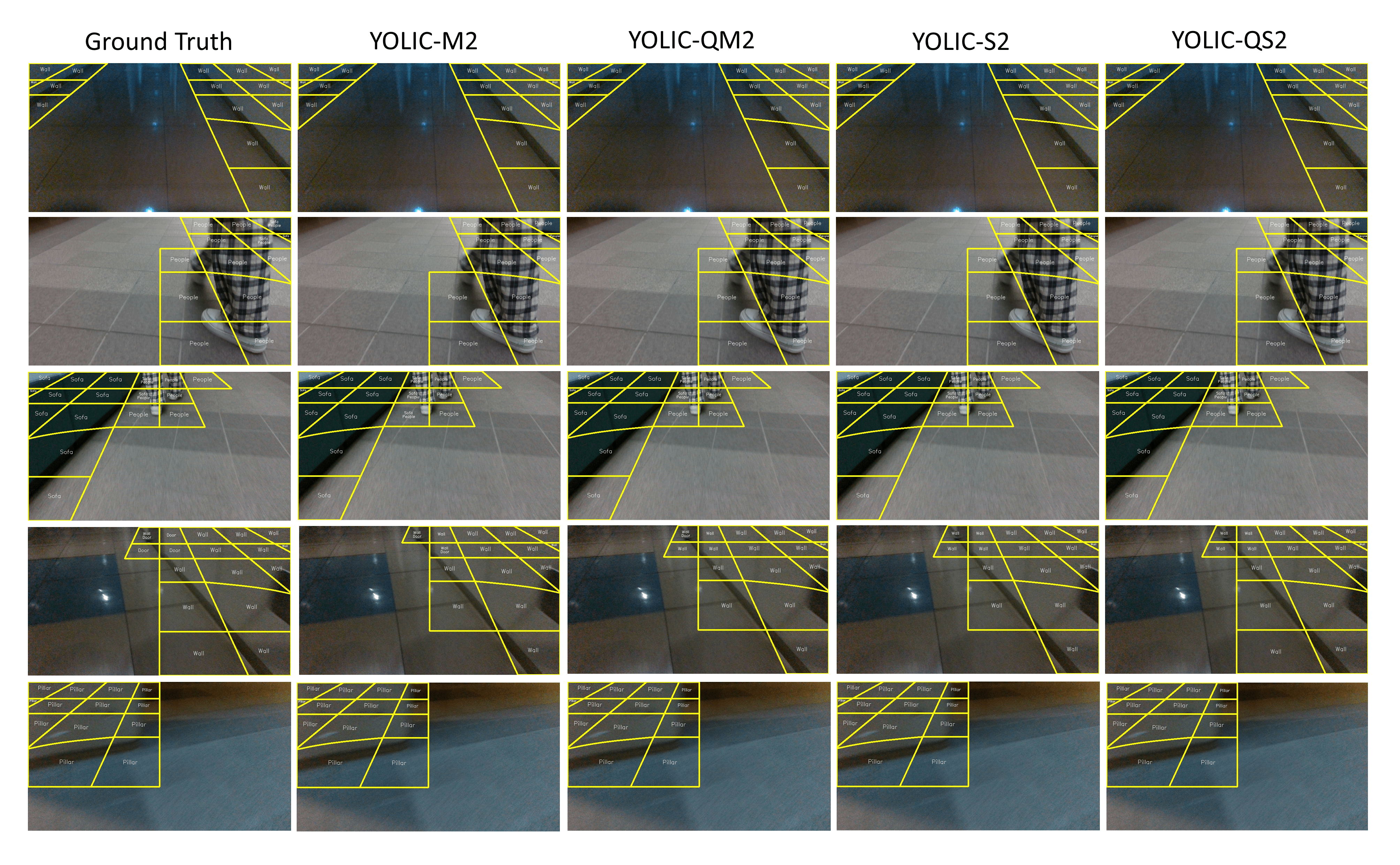}}
	\caption{This figure displays five cases of indoor obstacle detection, each with the corresponding output from four distinct YOLIC models.}
	\label{indoor}
\end{figure*}

To evaluate the effectiveness of this application, we curated a dataset consisting of 6410 distinct video frames. These frames were selected at one-second intervals from multiple videos to ensure no two frames were identical. The videos encapsulate a variety of indoor environments teeming with diverse obstacles. Six distinct categories of objects - ``Sofa", ``Pillar", ``Door", ``Wall", ``People", and ``Other", were marked as the focal points for this experiment. Consequently, the output of the YOLIC application for this study is calculated to be $30\times(6+1)=210$. The dataset was subsequently distributed randomly into different phases of our experiment: 70\% of the images were set aside for training, 10\% for validation, and the residual 20\% for the testing phase.

The indoor obstacle detection experiment results are presented in Table \ref{t6}, which demonstrates the successful detection and localization of various indoor obstacles using irregularly shaped CoIs. Due to the specificities of this cell configuration, we refrained from drawing comparisons with conventional YOLO series algorithms, which are not designed to handle such irregular CoIs. In addition to all category comparison, we also assessed the binary detection capabilities for each CoI. As summarized in Table \ref{t7}, all YOLIC models performed admirably in distinguishing between ``obstacle" and ``road" categories, further validating the robustness of our method. Figure \ref{indoor} showcases a selection of detection results, illustrating the YOLIC's ability to accurately identify and localize objects of interest within the defined CoIs in indoor environments.

\begin{table*}
\caption{Performance comparison of different YOLIC model in terms of precision and recall for each obstacle category in the indoor environment experiment. }
\label{t6}
\centering
\resizebox{\linewidth}{!}{%
\begin{tblr}{
  cells = {c},
  cell{1}{1} = {r=2}{},
  cell{1}{2} = {r=2}{},
  cell{1}{3} = {c=3}{},
  cell{1}{6} = {c=3}{},
  cell{1}{9} = {c=3}{},
  cell{1}{12} = {c=3}{},
  hline{1,3,7} = {-}{},
}
Method    & Input Size & Sofa      &        &          & Wall      &        &          & Pillar    &        &          & People    &        &          \\
          &            & Precision & Recall & F1-score & Precision & Recall & F1-score & Precision & Recall & F1-score & Precision & Recall & F1-score \\
YOLIC-M2  & 224 × 224    & 0.9613    & 0.9309 & 0.9459   & 0.9642    & 0.9597 & 0.9619   & 0.9346    & 0.9115 & 0.9229   & 0.9242    & 0.9314 & 0.9278   \\
YOLIC-S2  & 224 × 224    & 0.9452    & 0.9200 & 0.9324   & 0.9614    & 0.9511 & 0.9562   & 0.9135    & 0.8992 & 0.9063   & 0.9302    & 0.9052 & 0.9175   \\
YOLIC-QM2 & 224 × 224    & 0.9653    & 0.8776 & 0.9194   & 0.9660    & 0.9495 & 0.9577   & 0.9433    & 0.8757 & 0.9082   & 0.9304    & 0.9080 & 0.9191   \\
YOLIC-QS2 & 224 × 224    & 0.9241    & 0.9013 & 0.9126   & 0.9533    & 0.9479 & 0.9506   & 0.8847    & 0.9104 & 0.8974   & 0.9254    & 0.8960 & 0.9105   \\
          &            &           &        &          &           &        &          &           &        &          &           &        &          
\end{tblr}
}
\centering
\begin{tblr}{
  cells = {c},
  cell{1}{1} = {r=2}{},
  cell{1}{2} = {r=2}{},
  cell{1}{3} = {c=3}{},
  cell{1}{6} = {c=3}{},
  cell{1}{9} = {c=3}{},
  hline{1,3,7} = {-}{},
}
Method    & Input Size & Door      &        &          & Other     &        &          & All       &        &           \\
          &            & Precision & Recall & F1-score & Precision & Recall & F1-score & Precision & Recall & F1-score  \\
YOLIC-M2  & 224 × 224    & 0.9412    & 0.9418 & 0.9415   & 0.9416    & 0.8753 & 0.9072   & 0.9445    & 0.9251 & 0.9347  \\
YOLIC-S2  & 224 × 224    & 0.9403    & 0.9165 & 0.9282   & 0.9398    & 0.8475 & 0.8913   & 0.9384    & 0.9066 & 0.9222  \\
YOLIC-QM2 & 224 × 224    & 0.9451    & 0.9270 & 0.9360   & 0.9408    & 0.8514 & 0.8939   & 0.9485    & 0.8982 & 0.9227   \\
YOLIC-QS2 & 224 × 224    & 0.9311    & 0.8868 & 0.9084   & 0.9307    & 0.8243 & 0.8743   & 0.9249    & 0.8945 & 0.9094 
\end{tblr}
\end{table*}

\begin{table*}
\centering
\caption{Performance comparison of different YOLIC model for binary detection in the indoor environment experiment.}
\label{t7}
\begin{tblr}{
		cells = {c},
		cell{1}{1} = {r=2}{},
		cell{1}{2} = {r=2}{},
		cell{1}{3} = {c=3}{},
		cell{1}{6} = {c=3}{},
		cell{1}{9} = {c=3}{},
		hline{1,3,7} = {-}{},
	}
	Method    & Input Size & Risk      &        &          & Road      &        &          & All       &        &          \\
	&            & Precision & Recall & F1-score & Precision & Recall & F1-score & Precision & Recall & F1-score \\
	YOLIC-M2  & 224 × 224        & 0.9699    & 0.9703 & 0.9701   & 0.9854    & 0.9852 & 0.9853   & 0.9777    & 0.9778 & 0.9777   \\
	YOLIC-S2  & 224 × 224        & 0.9678    & 0.9601 & 0.9639   & 0.9806    & 0.9844 & 0.9825   & 0.9742    & 0.9723 & 0.9732   \\
	YOLIC-QM2 & 224 × 224        & 0.9683    & 0.9654 & 0.9668   & 0.9831    & 0.9845 & 0.9838   & 0.9757    & 0.9750 & 0.9753   \\
	YOLIC-QS2 & 224 × 224        & 0.9540    & 0.9661 & 0.9600   & 0.9833    & 0.9772 & 0.9802   & 0.9687    & 0.9716 & 0.9701   
\end{tblr}
\end{table*}

\subsection{Performance Validation on Public Cityscapes Dataset}
The Cityscapes dataset\cite{Cordts_2016_CVPR} is a large-scale dataset for diverse urban street scenes. The dataset contains 5,000 images with fine annotations, divided into 2,975 training, 500 validation, and 1,525 test images. Given the fact that the Cityscapes dataset does not publicly provide annotations for its test images, it is a common practice to conduct both training and validation on the aforementioned training set in our experiments. The validation set, therefore, is reserved solely for the final comparison of results.

\begin{figure}[htbp]
	\centerline{\includegraphics[scale=0.65]{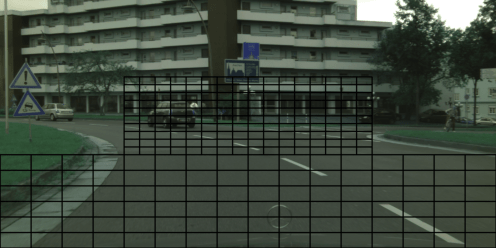}}
	\caption{Cell configuration for the Cityscapes dataset experiment. A total of 256 CoIs are employed, with 160 smaller CoIs focused on the central region of the image for more precise object localization at a distance, and 96 larger CoIs positioned in the lower region to cover closer objects. The arrangement is designed to prioritize the detection of people, vehicles, and road backgrounds in front of the vehicle.}
	\label{fig3}
\end{figure}

In applying YOLIC to the Cityscapes dataset, our attention was particularly drawn to a subset of object categories that are of substantial relevance to driving systems. These categories encompassed people, vehicles, and road backgrounds. We formulated a cell configuration that accounts for the spatial distribution and scale of objects commonly found in urban environments. As shown in Figure \ref{fig3}, we assigned a total of 256 CoIs to accentuate the detection of objects in the distant surroundings of a vehicle. Furthermore, we positioned 96 larger CoIs in front of the car to effectively cover objects that are closer and more likely to impact the vehicle's movement.

\begin{table*}
	\centering
	\caption{Comparative performance metrics for different YOLO and YOLIC models (using MobileNetV2 and ShuffleNetV2, as well as their quantized versions) across various object categories. }
	\label{t1}
	\resizebox{\linewidth}{!}{%
\begin{tblr}{
		cells = {c},
		cell{1}{1} = {r=2}{},
		cell{1}{2} = {r=2}{},
		cell{1}{3} = {c=3}{},
		cell{1}{6} = {c=3}{},
		cell{1}{9} = {c=3}{},
		cell{1}{12} = {c=3}{},
		hline{1,3,12} = {-}{},
	}
	Method    & Input Size & Vehicle   &        &          & People    &        &          & Other     &        &          & All       &        &          \\
	&            & Precision & Recall & F1-score & Precision & Recall & F1-score & Precision & Recall & F1-score & Precision & Recall & F1-score \\
	YOLOv5-N  & 640 × 640        & 0.8730    & 0.8560 & 0.8644   & 0.7780    & 0.7800 & 0.7790   & 0.9190    & 0.6870 & 0.7862   & 0.8567    & 0.7743 & 0.8134   \\
	YOLOv5-S  & 640 × 640        & 0.8730    & 0.8770 & 0.8750   & 0.8400    & 0.7980 & 0.8185   & 0.9370    & 0.6710 & 0.7820   & 0.8833    & 0.7820 & 0.8296   \\
	YOLOv6-N  & 640 × 640        & 0.8250    & 0.5940 & 0.6907   & 0.7030    & 0.6350 & 0.6673   & 0.4620    & 0.4710 & 0.4665   & 0.6633    & 0.5667 & 0.6112   \\
	YOLOv8-N  & 640 × 640        & 0.8700    & 0.8700 & 0.8700   & 0.8200    & 0.7630 & 0.7905   & 0.9090    & 0.7390 & 0.8152   & 0.8663    & 0.7907 & 0.8268   \\
	YOLOv8-S  & 640 × 640        & 0.8840    & 0.8600 & 0.8718   & 0.8200    & 0.8390 & 0.8294   & 0.9420    & 0.6700 & 0.7831   & 0.8820    & 0.7897 & 0.8333   \\
	YOLIC-M2  & 224 × 224        & 0.8803    & 0.8394 & 0.8594   & 0.7451    & 0.6160 & 0.6744   & 0.9168    & 0.9286 & 0.9227   & 0.8474    & 0.7947 & 0.8202   \\
	YOLIC-S2  & 224 × 224        & 0.8495    & 0.8134 & 0.8311   & 0.6688    & 0.5505 & 0.6039   & 0.9004    & 0.9207 & 0.9104   & 0.8062    & 0.7615 & 0.7832   \\
	YOLIC-QM2 & 224 × 224        & 0.8804    & 0.7936 & 0.8347   & 0.7544    & 0.5185 & 0.6146   & 0.9091    & 0.9186 & 0.9138   & 0.8480    & 0.7436 & 0.7923   \\
	YOLIC-QS2 & 224 × 224        & 0.8595    & 0.7811 & 0.8184   & 0.7077    & 0.4983 & 0.5848   & 0.9111    & 0.8972 & 0.9041   & 0.8261    & 0.7255 & 0.7726   
\end{tblr}
}
\end{table*}

\begin{figure*}[htbp]
	\centerline{\includegraphics[scale=0.52]{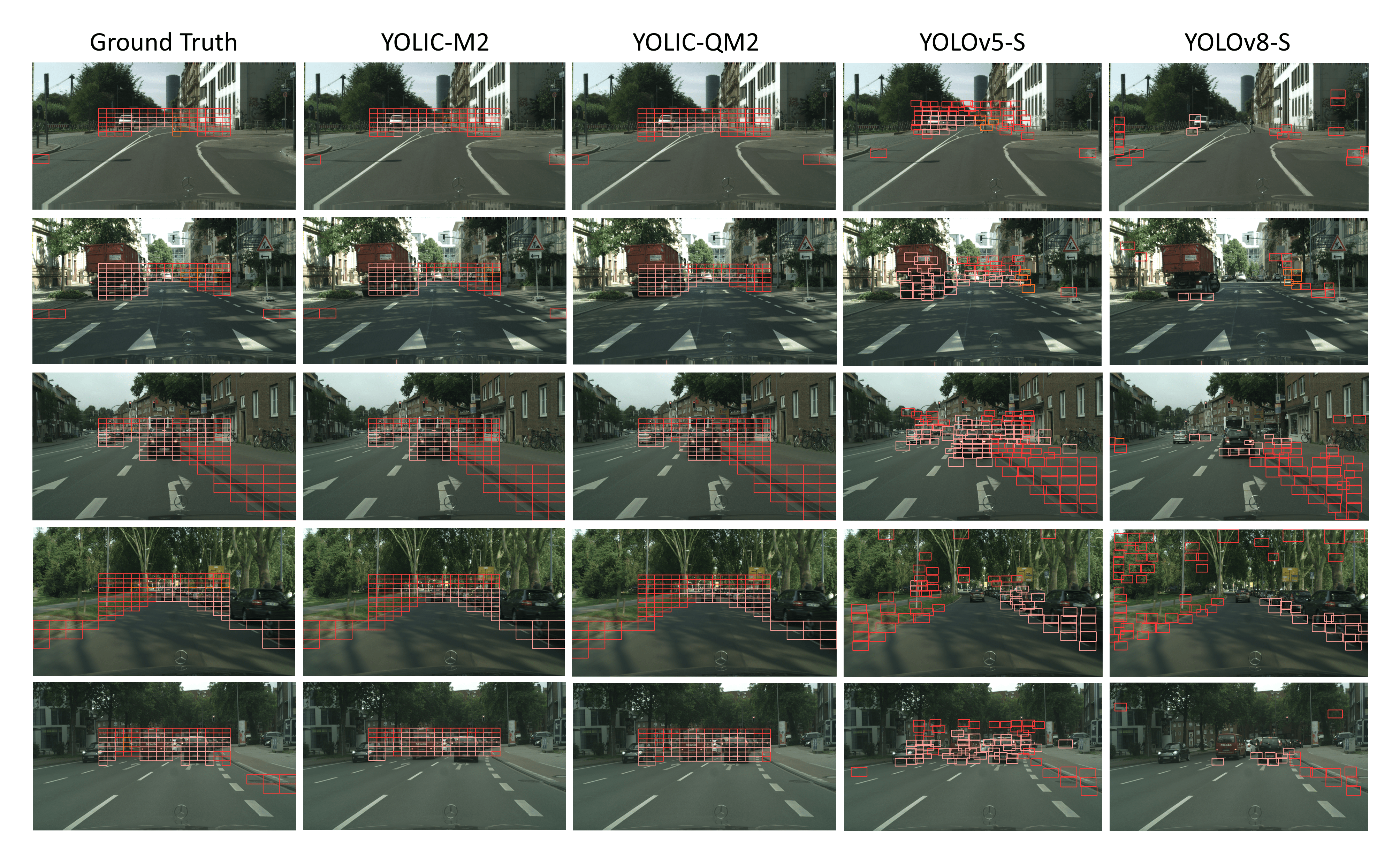}}
	\caption{We compare the detection results of our proposed models with the YOLO by visualizing the output on sample Cityscapes images. The four models with the highest F1-scores are selected for the comparison.}
	\label{cityscapesfig}
\end{figure*}

After establishing the cell configuration, we transformed the pixel-wise labeled images from the Cityscapes dataset into a YOLIC-compatible format. This involved assigning class labels to the respective CoIs, in accordance with our targeted object categories and the designed cell configuration. According to our experimental setup, the YOLIC network's output for the Cityscapes dataset consists of 256 CoIs and 4 categories, which include pedestrians, vehicles, other and road backgrounds, leading to a total of 1024 output values.

Table \ref{t1} provides a comprehensive comparison of the YOLIC method against other leading object detection models, demonstrating the substantial effectiveness of our YOLIC model. Notably, the YOLIC model, whether incorporating MobileNet V2 or ShuffleNet V2, consistently displays competitive performance. For instance, considering the category `Vehicle,' YOLIC-M2 demonstrates a precision of 0.8803 and a recall of 0.8394, both competitive numbers relative to other models. Further, the YOLIC-S2 model exhibits a precision of 0.8495 and a recall of 0.8134. Upon quantization, the models (YOLIC-QM2 and YOLIC-QS2) still maintain their robust performance. The YOLIC-QM2 model offers a precision of 0.8804 and a recall of 0.7936, while the YOLIC-QS2 version provides a precision of 0.8595 and a recall of 0.7811. When considering the overall performance across all categories, YOLIC models maintain a competitive edge. Specifically, YOLIC-M2 achieves an F1-score of 0.8202, which is comparable to the high-performing YOLO models. We also furnish a visual representation of the detection results in Figure \ref{cityscapesfig}. The figure encompasses a selection of sample images from the Cityscapes dataset, overlaid with corresponding outputs.

In addition to the evaluation of individual object categories, we also assess the binary classification for each CoI, with the results illustrated in Table \ref{t2}. In the `Road' category, the YOLIC models exhibit a highly accurate and consistent performance. The YOLIC-M2 model showcases a strong ability in correctly identifying road elements with a precision of 0.9704, while also effectively capturing most of these elements with a recall of 0.9585. Similarly, the YOLIC-S2 model demonstrates a commendable balance between precision and recall, indicating a reliable detection capability. The quantized versions of these models, while designed for computational efficiency, still retain a high level of accuracy and completeness in their detections. YOLIC-QM2 and YOLIC-QS2 models have precision and recall scores that are on par with the non-quantized models, showcasing the robustness of these versions despite their simplified structures. This binary classification results further underscore the efficacy of the YOLIC method in accurate object detection within predefined CoIs. 
\begin{table*}
	\centering
	\caption{Binary classification performance of the proposed YOLIC models for Cityscapes dataset.}
	\label{t2}
		\begin{tblr}{
				cells = {c},
				cell{1}{1} = {r=2}{},
				cell{1}{2} = {r=2}{},
				cell{1}{3} = {c=3}{},
				cell{1}{6} = {c=3}{},
				cell{1}{9} = {c=3}{},
				hline{1,3,7} = {-}{},
			}
			Method    & Input Size & Risk      &        &          & Road      &        &          & All       &        &          \\
			&            & Precision & Recall & F1-score & Precision & Recall & F1-score & Precision & Recall & F1-score \\
			YOLIC-M2  & 224 × 224        & 0.9601    & 0.9716 & 0.9658   & 0.9704    & 0.9585 & 0.9644   & 0.9653    & 0.9651 & 0.9651   \\
			YOLIC-S2  & 224 × 224        & 0.9505    & 0.9669 & 0.9586   & 0.9653    & 0.9482 & 0.9567   & 0.9579    & 0.9576 & 0.9577   \\
			YOLIC-QM2 & 224 × 224        & 0.9517    & 0.9662 & 0.9589   & 0.9647    & 0.9496 & 0.9571   & 0.9582    & 0.9579 & 0.9580   \\
			YOLIC-QS2 & 224 × 224        & 0.9475    & 0.9664 & 0.9569   & 0.9647    & 0.9449 & 0.9547   & 0.9561    & 0.9557 & 0.9559   
		\end{tblr}
\end{table*}

\subsection{Speed Comparison with State-of-the-art Systems}
In this subsection, we turn our attention to the speed comparison of YOLIC with other state-of-the-art lightweight object detection algorithms, with an emphasis on the suitability for real-time detection in resource-constrained environments, such as IoT edge devices. For this purpose, we utilized the Raspberry Pi 4B as our testing platform, given its widespread use, representation of constrained environments, and the replicability of our experiments.

Two sets of tests were conducted to evaluate the speed performance of YOLIC, one with FP16 precision models and the other with quantized models. The results of these tests are presented in Table \ref{t8} and Table \ref{t9}, respectively. These tables provide a comprehensive comparison of YOLIC with other lightweight object detection algorithms. From Table 
\ref{t8}, YOLIC-S2 with an input size of 224 × 224 achieved a notable FPS of 34.01, considerably outperforming other algorithms. In Table \ref{t9}, the speed superiority of YOLIC is further exemplified in quantized models. When the YOLIC-M2 model is quantized and run on the PyTorch framework, the FPS jumps to 35.72. The quantized version of YOLIC-S2 running on the ncnn framework, maintains a high FPS of 40.06.

\begin{table}
\centering
\caption{Speed comparison of YOLIC with other state-of-the-art lightweight object detection algorithms using FP16 precision models. }
\label{t8}
\resizebox{\linewidth}{!}{%
\begin{tblr}{
  cells = {c},
  hline{1-2,14} = {-}{},
}
Method         & Input Size & FPS   & Post-Processing & \#Params & FLOPs \\
YOLOv5-N       & 640 × 640        & 1.89  & Yes             & 1.9M     & 4.5G  \\
YOLOv5-S       & 640 × 640        & 0.92   & Yes             & 7.2M     & 16.5G \\
YOLOv6-N       & 640 × 640        & 2.00  & Yes             & 4.7M     & 11.4G \\
YOLOv7-T       & 640 × 640        & 1.72  & Yes             & 6.2M     & 13.7G \\
YOLOv7-T       & 416 × 416        & 4.34  & Yes             & 6.2M     & 5.8G  \\
YOLOv8-N       & 640 × 640        & 2.00  & Yes             & 3.2M     & 8.7G  \\
YOLOv8-S       & 640 × 640        & 0.86   & Yes             & 11.2M    & 28.6G \\
NanoDet-m      & 320 × 320        & 12.24 & Yes             & 0.95M    & 0.72G \\
NanoDet-Plus-m & 416 × 416        & 5.35  & Yes             & 1.17M    & 0.90G \\
MobileNet-SSD  & 300 × 300        & 7.28  & Yes             & 6.8M     & 2.50G \\
YOLIC-M2       & 224 × 224        & 13.48 & No              & 3.5M     & 0.62G \\
YOLIC-S2       & 224 × 224        & 34.01 & No              & 2.28M    & 0.30G 
\end{tblr}
}
\end{table}

\begin{table}
\centering
\caption{Speed comparison of YOLIC with other state-of-the-art lightweight object detection algorithms using quantized models. The metrics are the same as in Table \ref{t8}, but additional columns indicate the frameworks used for running the quantized models.}
\label{t9}
\resizebox{\linewidth}{!}{%
\begin{tblr}{
  cells = {c},
  hline{1-2,15} = {-}{},
}
Method         & framework & Input Size & FPS   & Post-Processing & \#Params & FLOPs \\
YOLOv5-N       & ncnn      & 640 × 640        & 2.81  & Yes             & 1.9M     & 4.5G  \\
YOLOv5-S       & ncnn      & 640 × 640        & 1.38   & Yes             & 7.2M     & 16.5G \\
YOLOv6-N       & ncnn      & 640 × 640        & 3.63  & Yes             & 4.7M     & 11.4G \\
YOLOv7-T       & ncnn      & 640 × 640        & 3.81  & Yes             & 6.2M     & 13.7G \\
YOLOv7-T       & ncnn      & 416 × 416        & 6.40  & Yes             & 6.2M     & 5.8G  \\
YOLOv8-N       & ncnn      & 640 × 640        & 2.75  & Yes             & 3.2M     & 8.7G  \\
YOLOv8-S       & ncnn      & 640 × 640        & 1.25   & Yes             & 11.2M    & 28.6G \\
NanoDet-m      & ncnn      & 320 × 320        & 15.59 & Yes             & 0.95M    & 0.72G \\
NanoDet-Plus-m & ncnn      & 416 × 416        & 6.68  & Yes             & 1.17M    & 0.90G \\
MobileNet-SSD  & ncnn      & 300 × 300       & 12.87 & Yes             & 6.8M     & 2.50G \\
YOLIC-QM2       & ncnn      & 224 × 224        & 28.42 & No              & 3.5M     & 0.62G \\
YOLIC-QM2       & Pytorch   & 224 × 224        & 35.72 & No              & 3.5M     & 0.62G \\
YOLIC-QS2       & ncnn      & 224 × 224        & 40.06 & No              & 2.28M    & 0.30G 
\end{tblr}
}
\end{table}
\subsection{Discussions}
In this section, we have showcased the efficacy and versatility of the YOLIC object detection methodology through various scenarios. A range of comprehensive experiments demonstrates the flexibility, efficiency, and adaptability of YOLIC.

In the outdoor risk detection experiment designed for low-cost electric scooters, we demonstrated that our proposed YOLIC methodology empowers resource-constrained IoT edge devices with the capability to conduct real-time object detection. Moreover, the indoor obstacle detection experiment featured a unique cell configuration with irregular shapes, highlighted YOLIC's versatility in adjusting to diverse CoI configurations. Lastly, through the urban scene interpretation experiment using the Cityscapes dataset, we affirmed YOLIC's robustness and reliability in handling complex urban scenarios.

Among the various models analyzed, YOLIC-S2 stands out. It is the only model capable of real-time detection under FP16 precision on Raspberry Pi. Its detection speed even exceeds 40 FPS under INT-8 precision. This model's outstanding speed and satisfactory performance render it the recommended choice for tasks requiring fast detection, such as outdoor risk detection and indoor obstacle avoidance. However, for tasks that require high-precision detection, we propose the YOLIC-M2 model. This model's performance on the Cityscapes dataset was on par with other YOLO comparative models, with only minor differences. Remarkably, this model also can achieve real-time detection at over 30 FPS under INT-8 precision.

\begin{figure*}[htbp]
	\centerline{\includegraphics[scale=0.47]{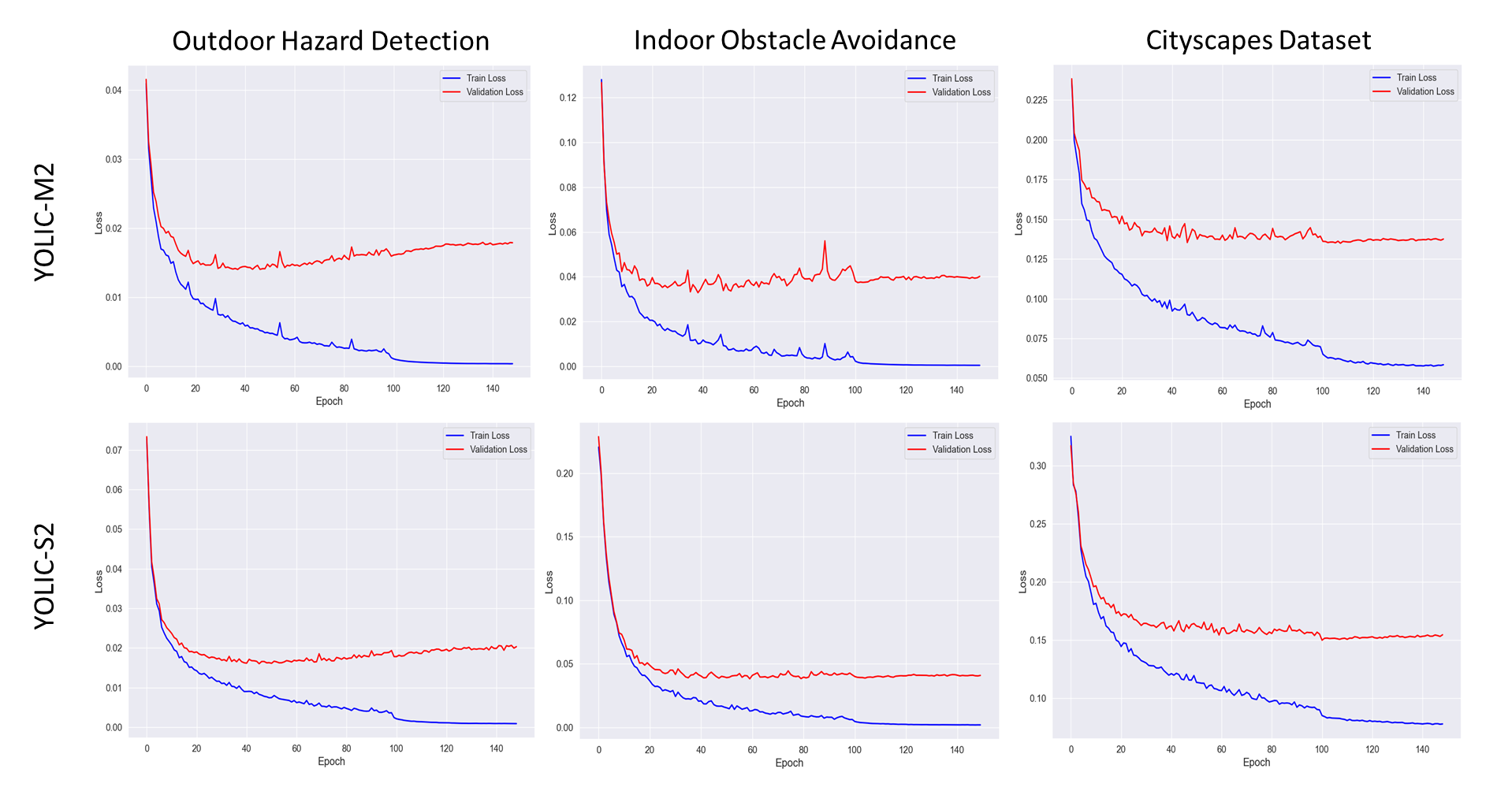}}
	\caption{Training loss trajectories for the YOLIC-M2 and YOLIC-S2 models across three different datasets.}
	\label{figloss}
\end{figure*}

During the training phase, as illustrated in Figure \ref{figloss}, both YOLIC-M2 and YOLIC-S2 demonstrated consistent loss reduction across all three datasets, validating the effectiveness of our training process and the adaptability of these models across varied datasets.

In terms of detection speed, YOLIC demonstrated superior efficiency in both FP16 precision and quantized models, especially on resource-constrained device Raspberry Pi 4B. Notably, YOLIC's competitive advantage in detection speed does not require additional post-processing. In addition, we can further increase the detection accuracy by leveraging consecutive video frames in practical applications.

\section{Conclusion}
In this paper, we have presented the YOLIC method, a unique approach to lightweight object localization and classification designed for edge devices. YOLIC employs a segmentation strategy based on predefined cells of interest and utilizes multi-label classification techniques. This effective combination allows YOLIC to recognize multiple objects within each CoI, thus making it adaptable to a wide array of detection scenarios. Our comprehensive experiments demonstrate YOLIC's ability to handle diverse and context-specific challenges. These range from complex urban scene interpretation to outdoor hazard detection for electric scooters, and indoor obstacle avoidance. Through the use of customized cell configurations, YOLIC maintains a high standard of detection accuracy while ensuring impressive detection speed. This ability to provide real-time detection results makes YOLIC stand out among current object detection methodologies. While YOLIC has demonstrated compelling results, there still remain some challenges that must be overcome. The increase in the number of CoIs can lead to a rise in the classification head parameters, potentially increasing computational complexity and memory usage. Our future work will aim to manage this trade-off, refining YOLIC's design further and enhancing its applicability across more scenarios.

\section*{Acknowledgments}

We would like to extend our heartfelt thanks to Mr. Huitao Wang for his invaluable assistance in the data collection and annotation processes. In addition, we are immensely grateful to the collaborative research company, which provided the equipment necessary for conducting our experiments.

\bibliographystyle{IEEEtran}
\bibliography{ref}

\begin{IEEEbiography}[{\includegraphics[width=1in,height=1.25in,clip,keepaspectratio]{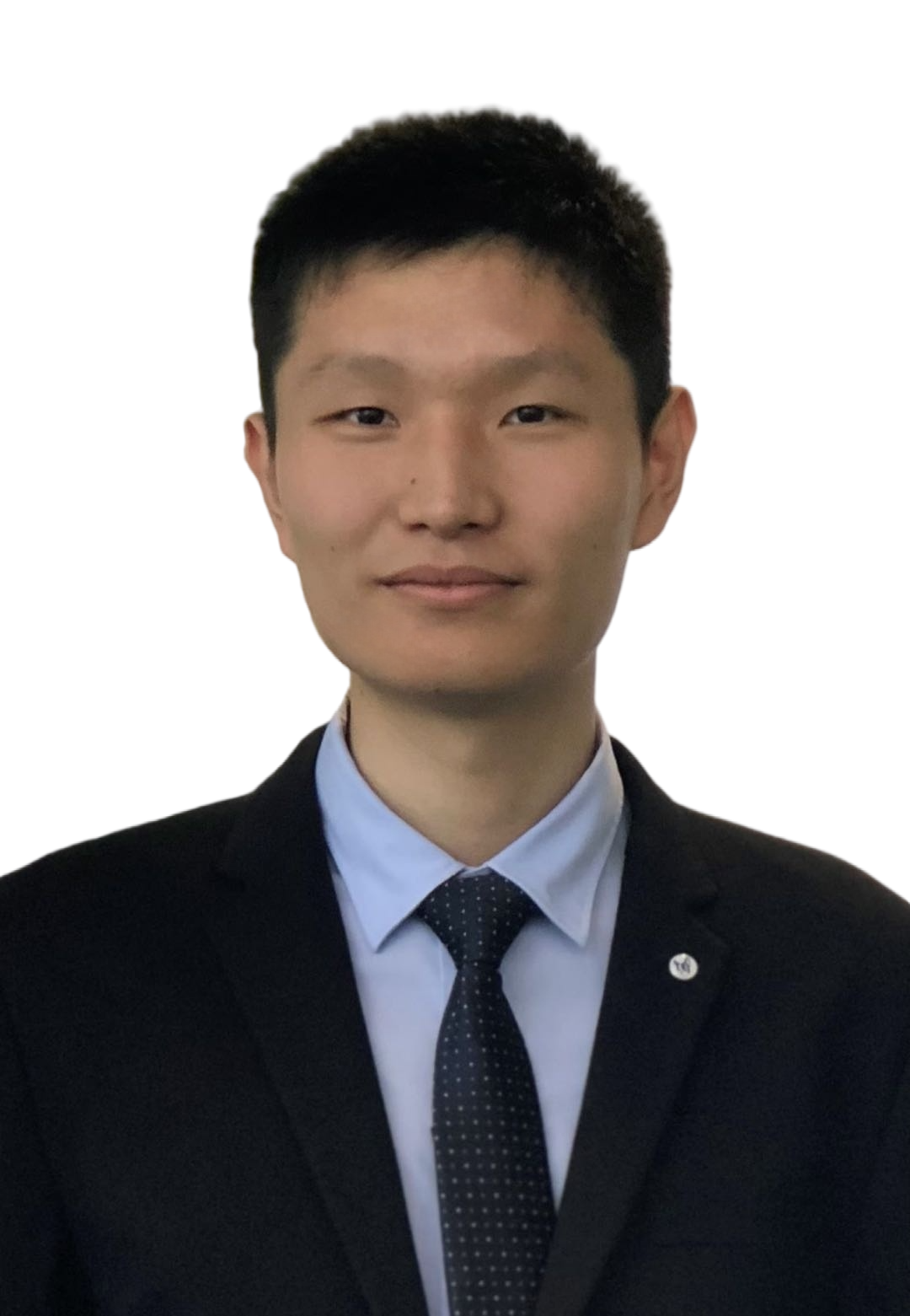}}]{KAI SU}
(Graduate Student Member, IEEE) received the B.Sc. degree in computer science from The University of Aizu, Japan, in 2019. He received his M.Sc. degree in computer science and engineering from The University of Aizu in 2021. He is currently pursuing a Ph.D. degree with the Graduate Department of Computer and Information Systems. His current research interests include computer vision, machine learning and their applications.
\end{IEEEbiography}

\begin{IEEEbiography}[{\includegraphics[width=1in,height=1.25in,clip,keepaspectratio]{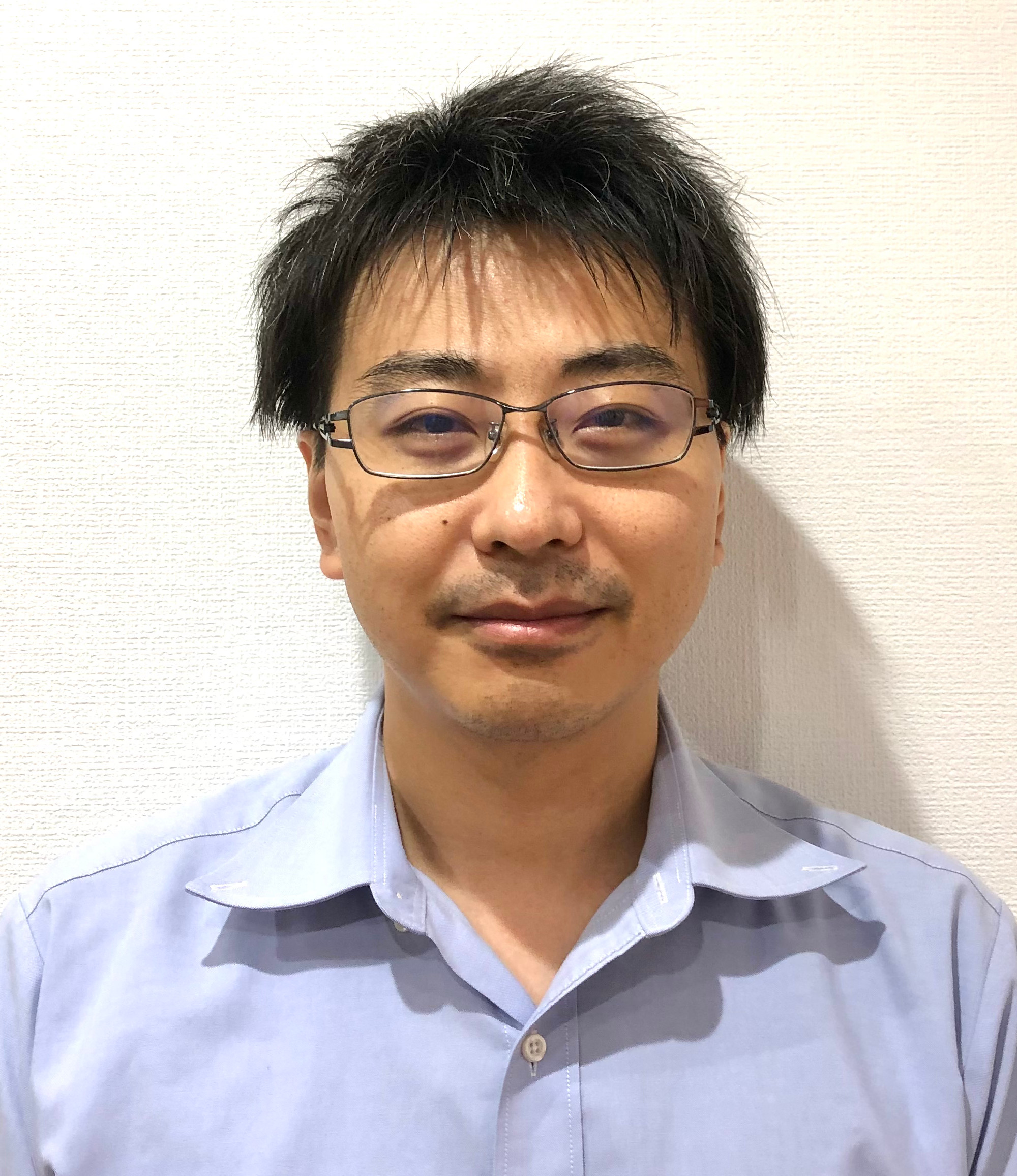}}]{YOICHI TOMIOKA}
(Member, IEEE) received the B.E., M.E., and D.E. degrees from the Tokyo Institute of Technology, Tokyo, Japan, in 2005, 2006, and 2009, respectively. He was a Research Associate with the Tokyo Institute of Technology, until 2009. He was an Assistant Professor with the Division of Advanced Electrical and Electronics Engineering, Tokyo University of Agriculture and Technology, until 2015. He was an Associate Professor with the School of Computer Science and Engineering, The University of Aizu, until 2018, where he has been a Senior Associate Professor, since 2019. His research interests include image processing, hardware acceleration, high-performance computing, electrical design automation, and combinational algorithms.
\end{IEEEbiography}

\begin{IEEEbiography}[{\includegraphics[width=1in,height=1.25in,clip,keepaspectratio]{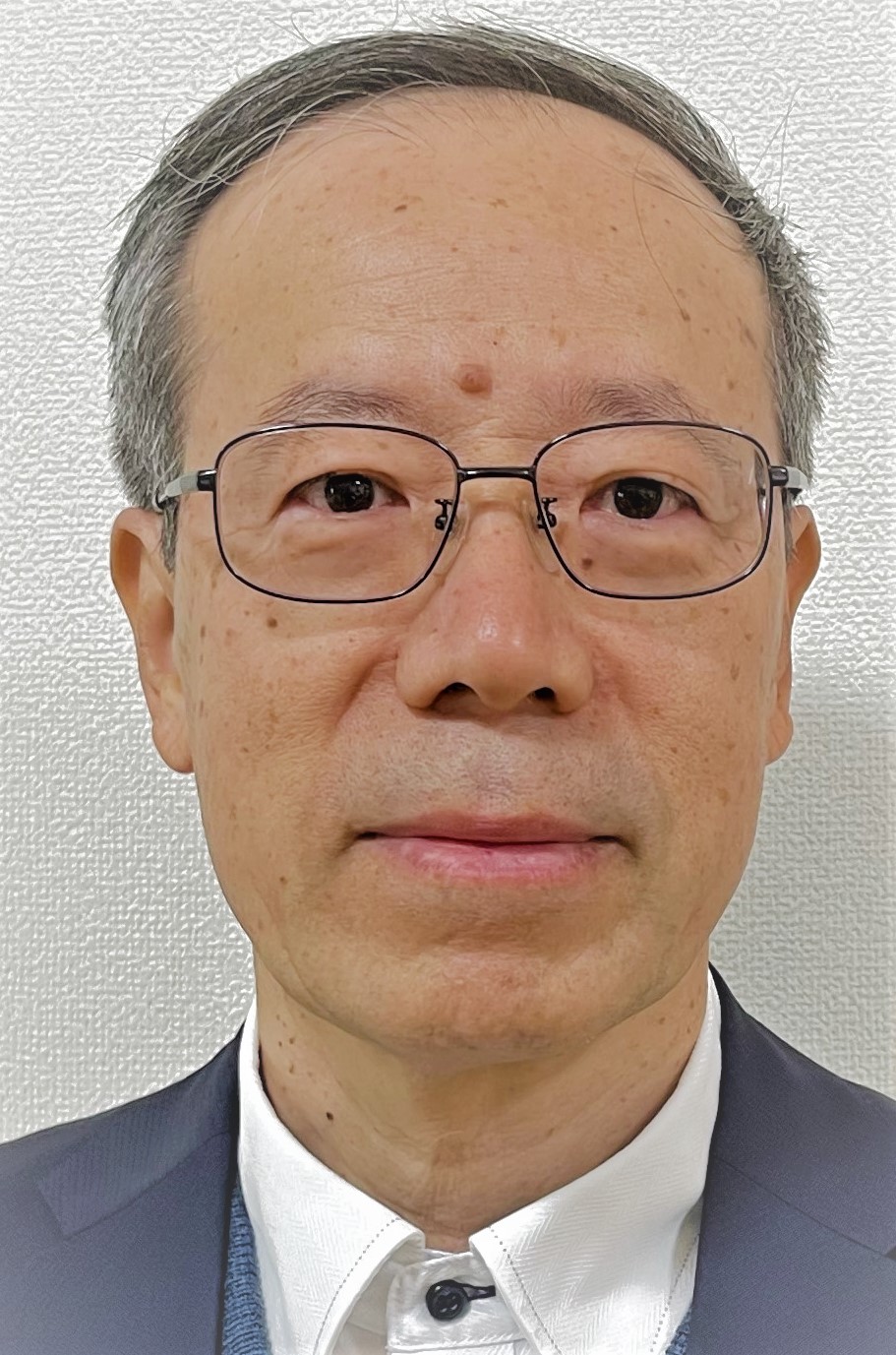}}]{Qiangfu Zhao}
(Senior Member, IEEE) received the Ph.D. degree from Tohoku University, Japan, in 1988. He joined the Department of Electronic Engineering, Beijing Institute of Technology, China, in 1988, first as a Postdoctoral Fellow and then an Associate Professor. Since October 1993, he has been an Associate Professor with the Department of Electronic Engineering, Tohoku University, Japan. In April 1995, he joined The University of Aizu, as an Associate Professor, where he became a tenure Full Professor, in April 1999. His research interests include image processing, pattern recognition, machine learning, and awareness computing.
\end{IEEEbiography}

\begin{IEEEbiography}[{\includegraphics[width=1in,height=1.25in,clip,keepaspectratio]{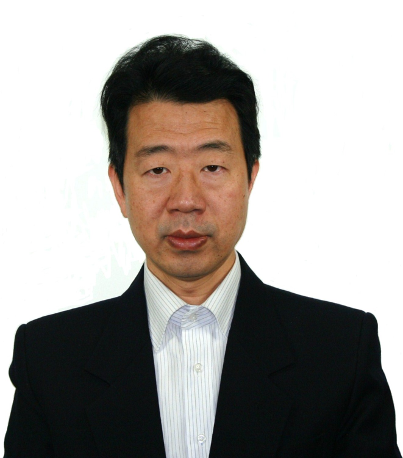}}]{Yong Liu}
is currently a professor at the University of Aizu, Japan. He was a guest professor in School of Computer Science, the China University of Geosciences, China, 2010. He was a researcher fellow at AIST Tsukuba Central 2, National Institute of Advanced Industrial Science and Technology, Japan, in 1999, and a lecturer in the State Key Laboratory of Software Engineering, Wuhan University, in 1994. His research interests include evolutionary computation and neural networks.
\end{IEEEbiography}

\vfill

\end{document}